\pdfoutput=1
\documentclass{article}

\usepackage[final,nonatbib]{nips_2016}

\usepackage[utf8]{inputenc} 
\usepackage[T1]{fontenc}    
\usepackage{url}            
\usepackage{booktabs}       
\usepackage{amsfonts}       
\usepackage{nicefrac}       
\usepackage{microtype}      

\usepackage{times}
\usepackage{graphicx} 
\usepackage{subfigure} 
\usepackage{amsmath}
\usepackage{amssymb}
\usepackage{epstopdf}
\usepackage{url}
\usepackage{caption}
\usepackage{multirow}
\usepackage{epsfig}
\usepackage{hhline}

\usepackage{algorithm}
\usepackage{algorithmic}

\title{Learning Sparse, Distributed Representations using the Hebbian Principle}

\author{
  Aseem Wadhwa \\
  Electrical and Computer Engineering\\
	University of California \\
	Santa Barbara, CA 93106 USA \\
  \texttt{aseem@ece.ucsb.edu} \\
  \And
	Upamanyu Madhow \\
  Electrical and Computer Engineering\\
	University of California \\
	Santa Barbara, CA 93106 USA \\
  \texttt{madhow@ece.ucsb.edu} \\
}

\begin{document}

\maketitle

\begin{abstract}

 The ``fire together, wire together'' Hebbian model is a central principle for learning in neuroscience, but surprisingly, it has found limited applicability in modern machine learning. 
In this paper, we take a first step towards bridging this gap, by developing flavors of competitive Hebbian learning which produce sparse, distributed neural codes using online adaptation with minimal tuning.  We propose an unsupervised algorithm, termed Adaptive Hebbian Learning (AHL).  We illustrate the distributed nature of the learned representations via output entropy computations for synthetic data, and demonstrate superior
performance, compared to standard alternatives such as autoencoders, in training a deep convolutional net on standard image datasets.

\end{abstract}

\section{Introduction}

Neuroscientific research has provided useful {\it architectural} insights for machine learning: for instance, multi-layered processing and convolutional architectures are directly inspired by the arrangement of cells in cortex. However, its contribution in developing practical {\it learning} mechanisms (e.g., weight adaptation) has been limited, despite the large body of neuroscientific research exploring development of synaptic weights. The ``fire together, wire together'' Hebbian principle \cite{hebb1952organisation}, in which synaptic weights connecting a pair of highly activated pre- and post-synaptic neurons are strengthened, has been the centerpiece in several prominent neuroscientific models of learning developed over the decades, such as self-organizing maps \cite{kohonen1982self} and adaptive resonance theory (ART models) \cite{carpenter1988art}. More recently, it continues to be used in modeling the development of cortical maps \cite{zylberberg2011sparse,bednar2014hebbian}. To date, however, these ideas have not been translated into practical machine learning algorithms, although there is renewed interest in this goal; for example, recent work in \cite{brito2015nonlinearhebbian} argues for Hebbian learning as a universal principle for feature learning.

Prior computational models for Hebbian learning show that, by combining it with components such as competition and inhibition, a variety of well-known
representations can be obtained, such as clustering (K-means), Principal Component Analysis (PCA), and sparse coding \cite{grossberg1976adaptive,sanger1989optimal,foldiak1990forming}. These results show that Hebbian learning can yield
results consistent with those from optimizing well known cost functions.  Indeed, most prior work on unsupervised learning is based on minimization of 
an explicit cost function (e.g., {\it energy} in Boltzmann machines and reconstruction error in autoencoders) along with regularization. However, such loss functions are at best an imperfect proxy for obtaining desirable data representations, and are in any case subject to local minima.

In this paper, we take a drastically different approach: instead of optimizing a cost function, we augment Hebbian rules with neuro-plausible mechanisms that {\it directly enforce core properties} of the resulting neural code: sparsity, ``distributed-ness'' and decorrelation. Several works, spanning both neuroscience \cite{field1994goal,foldiak1990forming,olshausen2003principles} and machine learning \cite{agrawal2014analyzing} areas, have identified these core properties as indicators of efficient representations learned in the lower and middle layers of deep architectures.  

{\bf Contributions and Organization:} After discussing related work (Section \ref{sec:related}), we present our algorithm, termed Adaptive Hebbian Learning (AHL), in Section \ref{sec:ahl}. AHL features competition among post-synaptic (i.e., output) neurons as well as, optionally, among synapses associated with pre-synaptic (i.e., input) neurons. Using a synthetic dataset, we demonstrate that synaptic competition reduces correlation across output neurons, creating a sparse, distributed code, in terms of increasing output entropy while still allowing accurate reconstruction (Section \ref{sec:artificial_data}). We then use AHL as a basis for unsupervised, bottom-up layer-wise learning in a standard deep convolutional net and demonstrate (Section \ref{sec:ahl_results}) excellent classification results for standard image datasets (MNIST, NORB, CIFAR), using supervised training of a linear SVM as the final layer. Comparing with layered autoencoders, we find that AHL converges faster, and that the features it provides yield better classification performance for CIFAR, and comparable performance for MNIST and NORB. 

To the best of our knowledge, the method presented here is the first successful attempt to translate the Hebbian principle into practical
machine learning algorithms that give good results on standard datasets. While we use classification performance as an evaluation metric in comparing our scheme against other unsupervised approaches, we note that our goal here is not (yet) to compete with the fully supervised backpropagation-based deep networks that have been developed using many man-years of effort for these databases.  Rather, our objective is to motivate further study on Hebbian architectures, by demonstrating
their promise for unsupervised and (eventually) semi-supervised learning.

\section{Related Work} \label{sec:related}

Much of the prior work in neuroscience on Hebbian computation has focused on front-end learning (e.g., of features like those observed in V1 simple cells) aimed at bridging the gap between theoretical and experimental studies of the visual cortex \cite{zylberberg2011sparse,bednar2014hebbian}, or on very simple recognition tasks \cite{fukushima1980neocognitron}. Most of these models are complex, involving detailed spike timing or lateral connections, and hence are not well matched to machine learning applications, unlike the abstractions presented in this paper.  

AHL is similar to online clustering algorithms which have been studied extensively \cite{zhong2005efficient,banerjee2004frequency,chung1994fuzzy}, such as fuzzy clustering and soft winner-take-all. However, the focus of this prior literature is on solving the conventional clustering problem, whereas our goal is to use AHL as a building block for feature extraction in a deep architecture. A single layer of K-means with a large number of features ($\sim 4000$) was shown to be effective for feature extraction in \cite{coates2011analysis}. Multi-layer K-means followed by SVM has also been used in \cite{coates2011selecting,akbas2014framework}. In contrast to the heavily correlated K-means centers, 
 due to our introduction of synaptic competition we are able to produce sparse, distributed codes, similar to
the empirically observed characteristics of high-performing supervised deep nets \cite{agrawal2014analyzing}.

Existing unsupervised learning approaches such as sparse coding \cite{olshausen1996emergence} and autoencoders \cite{vincent2008extracting,goodfellow2009measuring} seek to optimize a cost function which combines reconstruction error with some form of regularization. In contrast, AHL does not have an explicit cost function. Rather, its mechanisms for weight update, neuron recruitment and pruning,
are designed to {\it directly} promote sparsity and decorrelation.

 Unsupervised feature learning was originally employed for initializing deep nets prior to fine-tuning \cite{bengio2007greedy}, but was soon observed not to offer substantial advantages over carefully scaled random initializations. Our goal here is to use classification performance
to establish Hebbian learning as a competitive alternative to existing unsupervised strategies, rather than to compare against the highly optimized,
purely supervised backprop-based deep nets. However, while the latter represent the current state of the art in classification,
deep generative modeling continues to be an active area of research \cite{kingma2014semi,maaloe2016auxiliary}, with the goal of reducing dependence on labels, and our results motivate further investigation into Hebbian algorithms as building blocks for semi-supervised learning.

\section{AHL: Unsupervised Hebbian Learning}
\label{sec:ahl}

Consider a single layer of $d$ pre-synaptic neurons connected to $K$ post-synaptic neurons, with activations ${\bf x}\in\mathbb{R}^d$ and ${\bf y}\in\mathbb{R}^K$ respectively. Let ${\bf w}_j\in \mathbb{R}^d$ denote the weights incident on neuron ${\bf y}_j$. Our goal is to learn the weight matrix ${\bf W}_{K\times d}$. The most commonly used form of Hebbian principle for training the weights is given by:
\begin{equation}
{\bf w}'_j(t+1) = {\bf w}_j(t)+\eta {\bf x}h({\bf y}_j),\;{\rm where} \;{\bf y}_j = f({\bf w}_j^T{\bf x})
\label{eq:hebb_update}
\end{equation}
\begin{equation}
{\bf w}_j(t+1)=\frac{{\bf w}'_j(t+1)}{\left \| {\bf w}'_j(t+1) \right \|_2}
\label{eq:w_normalize}
\end{equation}

where $\eta$ is a small constant (learning rate) and $h(),f()$ could be linear or non-linear monotonic functions that causes an increase in the strength of ${\bf w}_{i,j}$ when ${\bf x}_i$ and ${\bf y}_j$ are high, producing ``fire together, wire together'' behavior. Weight normalization (\ref{eq:w_normalize}) ensures that the weights do not explode while training, and implicitly creates competition between incoming synapses. In Hebbian literature, it is also a common practice to normalize the inputs, i.e. $\left \| {\bf x} \right \|_2=1$. Replacing the plus with a minus in Eq. \ref{eq:hebb_update} results in an anti-Hebbian update, which produces inhibitory behavior. When $h(z)=f(z)=z$, all $K$ weight vectors converge to the largest principal component of the training dataset $\{{\bf x}_i,i=1,..,N\}$ \cite{oja1982simplified}, while a simple modification produces convergence to the top $K$ eigenvectors \cite{sanger1989optimal}. When $h(z)={\rm sign}(z)$ and the weights afferent to only the highest activated neuron are updated, a strategy known as the WTA (winner take all), the resulting weights converge to cluster centers, and we get an online version of spherical K-means \cite{grossberg1976adaptive,zhong2005efficient}. 
Other modifications of Hebbian learning can lead to online solutions for other cost functions such as ICA, sparse coding etc \cite{brito2015nonlinearhebbian}, but that typically requires introduction of lateral connections, which slows down inference.

Weights learnt in feedforward Hebbian architectures are either orthogonal (PCA) or highly correlated (K-means). The representations (i.e. the activation pattern ${\bf y}$) they lead to do not display the following fundamental properties of neural codes, that are widely believed to be important for effective learning. \\
$\bullet$ {\bf Sparse, Distributed Code}: such a code arises when each post-synaptic neuron has a low probability to fire and the stimulus is forced to be encoded in the activity of a few neurons \cite{field1994goal}. This kind of code is a compromise between local (compact code) and totally distributed representations (when there is a single grandmother active neuron) \cite{foldiak1990forming}. It has been argued that sparse distributed codes disentangle the {\it causes} leading to {\it meaningful} representations and also present a pattern that is easier for higher stages of the system to model \cite{olshausen2003principles}. Interestingly, such codes appear naturally in supervised deep nets, even though these constraints are not explicitly enforced. The work in \cite{agrawal2014analyzing} presents several empirical studies discovering the sparse and distributed nature of codes in the middle layers of a deep backpropagation trained network. \\
$\bullet$ {\bf Decorrelated neurons}: decorrelation between activities of post-synaptic neurons is important to generate naturally sparse representations that are concise and cover the input space efficiently. Decorrelation forces neurons to learn different features and is an important component in most neuroscientific models \cite{foldiak1990forming,zylberberg2011sparse,bednar2014hebbian}. However, decorrelation is different from orthogonality: neurons still need to capture the ``suspicious coincidences'' that define objects \cite{barlow1987cerebral}. For example, neurons representing concepts such as ``white'', ``furry'', ``has tail'', which would all fire when say a patch of cat is presented as stimuli, will have correlated activity patterns. 

Adaptive Hebbian Learning (AHL) builds on the WTA Hebbian update rule, with $h(z)={\rm sign}(z)$ and a max rectified non-linearity for $f()$. It combines this framework with several ideas, including competition between outgoing weights from a pre-synaptic neuron, and adaptive creation and culling of neurons, to obtain sparse, distributed representations. The key features of AHL are as follows:\\
(1) {\bf Bias}: The biases are set such that the average activity of each neuron is maintained at a fixed low level $A_{\rm bias}$ (e.g., $A_{\rm bias} =0.2$ results in average sparsity of $80\%$). This approach has been adopted in several papers building biologically plausible models 
\cite{zylberberg2011sparse,bednar2014hebbian} which conform to
the observation that neurons tend to have low mean firing rates that span a small range of values. Sparse autoencoders \cite{goodfellow2009measuring} also include a penalty term for the mean firing rate of the neurons. \\
(2) {\bf Decorrelation via adaptive adding and pruning of neurons}: A new neuron is recruited when an input pattern is `far away' from the existing neurons (as measured by cosine similarity) and is not well represented thus far (as measured by the summation of total activity generated by it).  An existing neuron is pruned when it is too highly correlated with other neurons. This simple scheme gradually grows the number of post-synaptic neurons (starting from 0) , depending on the representational power needed to model the input data. It is worth noting that the idea of recruiting neurons
adaptively was proposed in Grossberg's ART2 model \cite{carpenter1987art2} decades ago, but our system
architecture and the specifics of neuron recruitment, as well as the inclusion of pruning, is different.\\
(3) {\bf Synaptic competition}: To further increase the distributed nature of the code, we allow for a soft-WTA strategy in which we update the weights of the top $K_w \geq 2$ winners. However, this can increase the correlation between neurons. 
To counter this, we introduce competition between the {\it outgoing} weights from a pre-synaptic neuron: not all weights for the $K_w$ neurons are updated, but only those that are {\it stronger}. For instance, for top two winners ${\bf y}_j$ and ${\bf y}_k$, if ${\bf w}_{ij}>{\bf w}_{ik}$ then only ${\bf w}_{ij}$ (the connection from ${\bf x}_i$ to ${\bf y}_j$) undergoes the Hebbian update. Such weight competition has been observed in biological studies \cite{sanes1999development}\cite{song2000competitive}. In addition to producing a more distributed code, it also
leads to increased sparsity in the weights, often seen in supervised deep nets, and also recently reported for marginalized denoising autoencoders in \cite{chen2014marginalized}. For $K_w\!=\!1$, AHL is similar to spherical K-means, but differs because of the bias, and adaptive adding and pruning.

The pseudo code (Matlab based) for the AHL algorithm is provided in Algorithm \ref{alg:ahl_alg}. Vectors are denoted in bold small letters and matrices in bold large letters.

\begin{algorithm}[tb]
\caption{Adaptive Hebbian Learning (AHL)} \label{alg:ahl_alg}
\begin{algorithmic}[0]
\STATE INPUT: ${\bf X} (N\times d)$ \COMMENT{rows normalized}
\STATE OUTPUT: $K,{\bf W} (K \times d), {\bf b} (K \times 1) $
\STATE PARAMETERS:
\STATE $\eta(=10^{-2})$: learning rate
\STATE $A_{\rm bias}(=0.2)$: average activity level of neurons
\STATE $A_T(\sim 0.5-5), \rho_T(=0.6)$: control adding of neurons
\STATE $\rho_U(=0.8)$: prune if correlation above this
\STATE $K_w(\sim 1-3)$: control level of competition
\STATE $E(\sim 15)$: epochs
\STATE INITIALIZE:\;${\bf W} = {\bf X}(1,:);\;{\bf b}= 0;\;{\bf r} = A_{\rm bias};\;{\bf C}=0;\;{\bf e}=0$
\STATE {\bf for} ${\rm loop}=1,...,E$ \; {\bf do}
\STATE \quad {\bf for} $n=1,2,...,N$ \; {\bf do}
\STATE \quad \quad ${\bf x} = {\bf X}(n,:)$ 
\STATE \quad \quad ${\bf a} = {\rm max}({\bf 0},{\bf W}{\bf x}^T-{\bf b})$ 
\STATE \quad \quad ADD NEURON:
\STATE \quad \quad {\bf if} $\sum_{i=1}^K {\bf a}_i<A_T$ \& ${\rm max} \;{\bf W}{\bf x}^T<\rho_T$: 
\STATE \quad \quad \quad ${\bf W}=[{\bf W};{\bf x}];\;{\bf b} = [{\bf b};0];\;{\bf r} = [{\bf r};A_{\rm bias}];\;{\bf C} = [{\bf C} \;{\bf 0};\; {\bf 0} \;0];\;{\bf e} = [{\bf e};0]$
\STATE \quad \quad UPDATE WEIGHTS:
\STATE \quad \quad $U \leftarrow \{$ set of indices of top $K_w$ values in $\bf a$ $\}$
\STATE \quad \quad {\bf for} $j=1,...,d$ \;{\bf do}
\STATE \quad \quad \quad $w_{j,{\rm max}}=\underset{k \in U}{\rm max}\{w_{jk}\},\;\;w_{j,{\rm min}}=\underset{k \in U}{\rm min}\{w_{jk}\}$
\STATE \quad \quad \quad {\bf if} \;$x_j>0\;\rightarrow$ \;\;$\forall k\in U \;\& \;w_{jk}\geq 0.9 w_{j,{\rm max}}$\;: \; $w_{jk}=w_{jk}+\eta x_j$
\STATE \quad \quad \quad {\bf if} \;$x_j<0\;\rightarrow$ \;\;$\forall k\in U \;\& \;w_{jk}\leq 0.9 w_{j,{\rm min}}$\;: \; $w_{jk}=w_{jk}+\eta x_j$
\STATE \quad \quad Normalize each row ${\bf W}(k,:)\;\forall k\in U$
\STATE \quad \quad UPDATE BIAS: 
\STATE \quad \quad ${\bf r} = 0.99{\bf r} + .01{\rm sign}({\bf a})$;\;${\bf b}={\bf b} + .01({\bf r}-A_{\rm bias})$
\STATE \quad \quad PRUNE: 
\STATE \quad \quad {\bf for} $j,k=1,..,K$ ($k>j$)\; {\bf do} \; ${\bf C}(j,k)=0.9999{\bf C}(j,k)+.0001{\bf a}_j {\bf a}_k$
\STATE \quad \quad {\bf for} $j = 1,..,K$ \; {\bf do}\; ${\bf e}_j=0.9999{\bf e}_j+.0001{\bf a}_j^2$
\STATE \quad \quad {\bf if} ${\rm mod}(n,5000)=0\;\rightarrow$\; ${\bf C}_n(j,k)=\frac{{\bf C}(j,k)}{\sqrt{{\bf e}_j}\sqrt{{\bf e}_k}},\;\forall\;j,k>j=1,..,K$
\STATE \quad \quad {\bf while} $j^*\!,k^*\!=\!{\rm argmax}\;\!{\bf C}_n(j,k)$ \& ${\bf C}_n(j^*,k^*)\!>\!\rho_U\;\rightarrow$\;remove $k^*$ neuron
\end{algorithmic} 
\end{algorithm}

{\bf Code details and Parameters}: A moving average filter is used for updating the bias and also for recording the normalized correlation of activities between neurons. In our simulations, normalization of weights is performed using a weight decay term that is added to the Hebbian update (we omitted it from the pseudocode for easier readability). The decay term can be derived in a manner similar to the one discussed in \cite{oja1982simplified}, it is useful for improving the computation speed. Note that pruning is done in a greedy fashion, starting with the highest correlated pair. The adding criterion using a total activity threshold is something for which we do not have a theoretical basis. However in practice it seems to be a useful heuristic for capturing how well a given input is explained by the current set of neurons, given that they are not highly correlated to each other. In our simulations we have observed that the weight values converge typically in 10-20 epochs. We set the number $E$ to 15 in all our simulations. 

While AHL has a number of parameters, their settings are intuitive and typically represent a direct property (e.g., maximum correlation allowed, mean firing rate etc). Indeed, the values shown in the beginning of the pseudocode are used across datasets and layers, except for $A_T$, which is varied 
in order to obtain a varying number of neurons across layers. The learning rate is also not changed, since we normalize the input data at each layer. Thus, AHL is easy to tune: we get excellent performance for the classification task, for example, without requiring fine-tuning via cross-validation. However, it would be interesting to follow up this work with detailed ablation studies to capture the effect of each parameter in AHL.

\section{Synthetic data: Activation Code Structure}
\label{sec:artificial_data}

Before reporting on AHL within a deep net operating on complex image data, we first derive fundamental insight into the kinds of codes it generates using synthetic data.
We use L2 reconstruction error as a measure of information preservation (even though AHL is not optimized for an L2 cost function), and show that the performance
is similar to, or better than, that of standard clustering.  We use output entropy as a measure of representation
power and distributed-ness, and show that AHL is significantly better than clustering.

Since our inputs are normalized, we draw the synthetic data from a mixture of von Mises-Fisher (vMF) distributions,
where vMF is the analog of a Gaussian distribution on the hypersphere (e.g., the spherical K-means algorithm can be derived by assuming a generating density which is a mixture of vMF \cite{banerjee2004frequency}). 
The vMF pdf on a $(d-1)$-dimensional hypersphere in $\mathbb{R}^d$ is given by $f({\bf x};{\bf \mu};\kappa) = Z_d(\kappa){\rm exp}(\kappa {\bf x}^T{\bf \mu})$, where $\kappa \geq 0$ and $\left \| {\bf \mu} \right \|_2=1$, and the normalization constant $Z_d(\kappa)=(\kappa^{d/2-1})/((2\pi)^{d/2}I_{d/2-1}(\kappa))$. 

Our synthetic data $\{{\bf x}_i\!\in \!\mathbb{R}^{30},\;\!i\!=\!1,..,5000\}$ is generated using a mixture of five vMF distributions, each generating 1000 IID samples in $\mathbb{R}^{30}$ using the sampling procedure described in \cite{hornik2013movmf}. The concentration parameter $\kappa$ plays the role of inverse variance. We generate our data with $\kappa = 50({\rm low}),\;100,\;150({\rm high})$. 
We create a few instances of the dataset, and for each, apply the AHL algorithm with $K_w\!=\!1$ (WTA) and $K_w\!=\!2$. We set $\rho_T=0.8,\;A_T=1,\;\eta=0.1,\;E=7$. 
Pruning is switched off, because we wish to observe the correlation between cluster centers. For comparison, we also run standard batch spherical K-means algorithm (SPKM) \cite{zhong2005efficient}, initialized with centers randomly drawn from the dataset. For each instantiation of the dataset, the number of clusters are set to be the same as the output of AHL (we denote the two cases by SPKM 1 and SPKM 2 for $K_w\!=\!1$, $K_w\!=\!2$ respectively). Once the weights/cluster centers are learnt, the activation code for each datapoint is given by ${\bf y}_{K\times 1}={\rm max}({\bf 0},C{\bf x}-{\bf b})$, where $C_{K\times 30}$ is the collection of cluster centers. The bias ${\bf b}$ is chosen such that each cluster is active on average half the time (i.e. a mean firing rate of $A_{\rm bias}=0.5$).

We develop quantitative insight into the codes produced by AHL using the following measures:\\
{\it Reconstruction error:} For standard clustering, this equals
the distortion between the inputs and the nearest cluster centers ($\sum_i \left \| {\bf x}_i - {\bf c}_{\rm closest}\right \|_2$).  
This does not work for a distributed code, hence we consider the error ($1-{\bf x}^T\hat{\bf x}$) between a datapoint and its best estimate given its activation code ($\hat{\bf x}_i|{\bf y}_i$).
$\hat{\bf x}$ can be obtained as the solution to the following optimization problem:
	\begin{equation}
	\hat{\bf x} = \underset{\bf x}{\rm argmin}\;\left \| {\bf y}^{I_1} - (C^{I_1}{\bf x}-{\bf b}^{I_1} )\right \|_2^2\;;\; {\rm subject}\;\;{\rm to}\;\;C^{I_0}{\bf x}\leq {\bf b}^{I_0},\;\left \| {\bf x} \right \|_2=1	
	\end{equation}
	where the set of indices where ${\bf y}$ is zero and non-zero are denoted by $I_0$ and $I_1$ respectively. While this is non-convex due to the
	L2 constraint, we are able to obtain good solutions using ``convex-concave'' sequence convex programming (SCP) \cite{yuille2003concave}. This involves solving a sequence of convex programs by replacing the second constraint by two affine constraints ${\bf x}^T{\bf x} \leq 1$ and $\frac{{\bf x}_{\rm prev}^T{\bf x}}{\left \| {\bf x}_{\rm prev} \right \|}\geq 0.99$. We initialize by solving the problem without the L2 constraint, and then normalizing.  We find that 2-4 iterations suffice for convergence.\\
{\it Output Entropy:} To evaluate this, we perform binary quantization of the activation codes, which gives a codebook of size $2^K$ for $K$ output neurons.
We then compute the empirical pmf of the $2^K$ possible binary codewords activated by the data points.\\
{\it Weight correlation:} We compute the $^KC_2$ correlations between all pairs of weight vectors and sort them.

\begin{figure}[h]
\centering
\subfigure[]{
\includegraphics[trim=70 10 64 0,clip,width=0.5\columnwidth,scale =0.6]{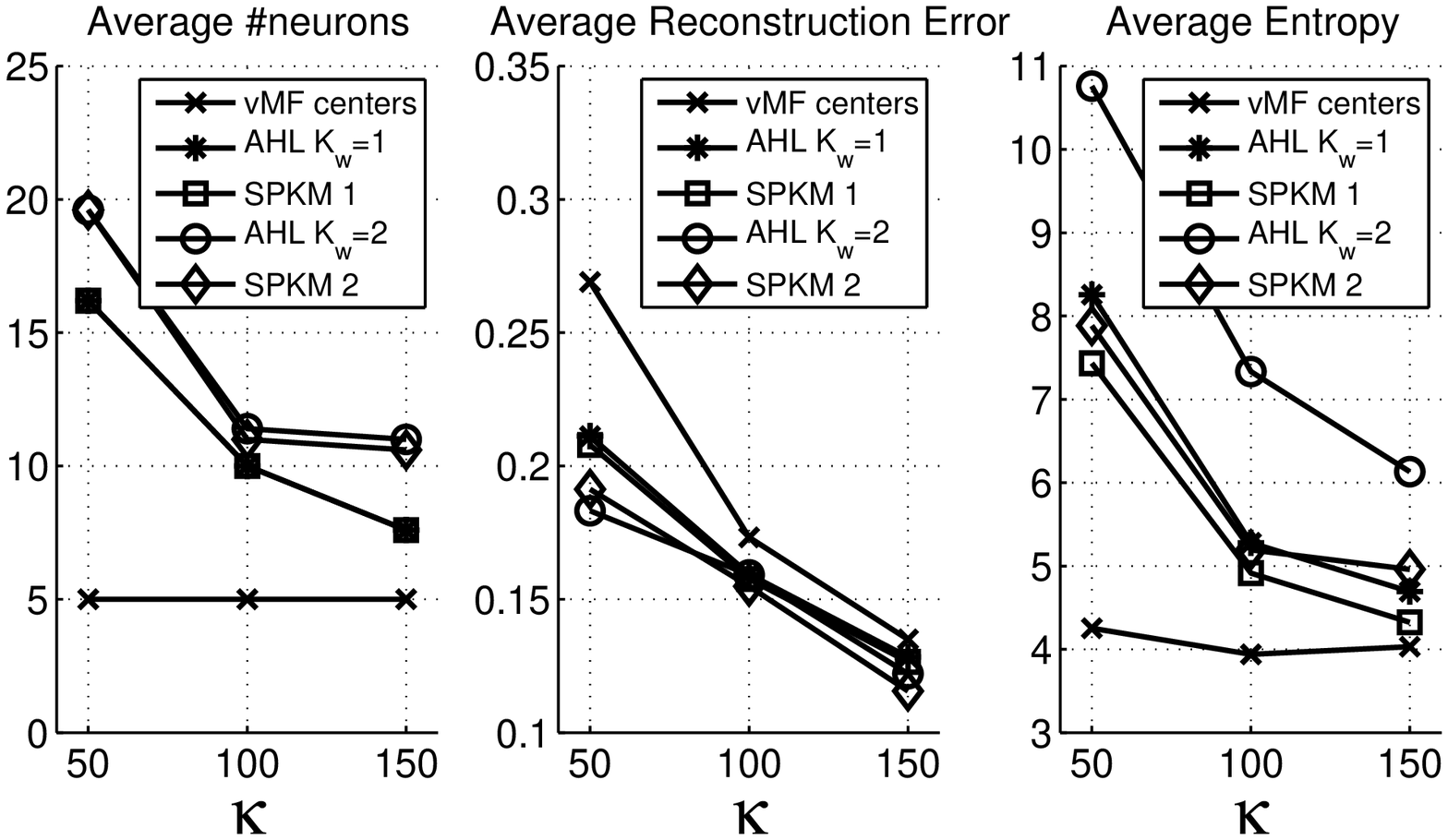}
\label{fig:syn_data_a}
}
\subfigure[]{\raisebox{8mm}{
\includegraphics[width=0.4\columnwidth,scale=0.6]{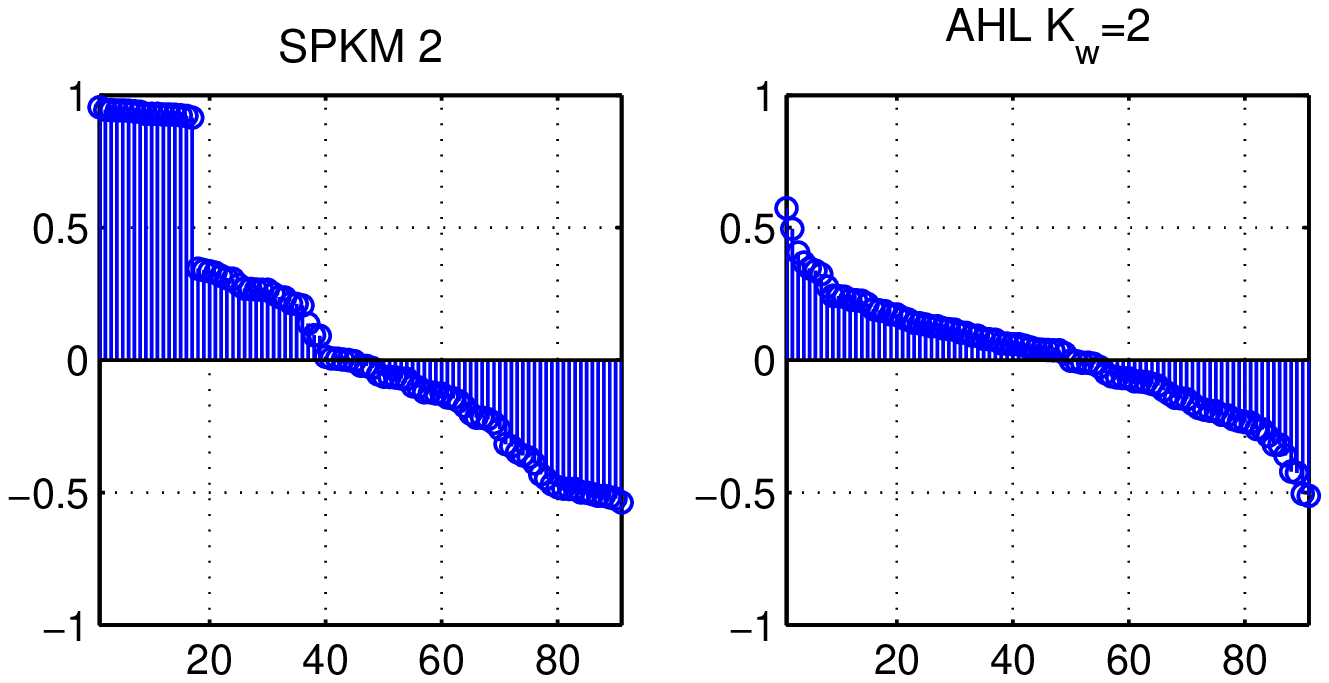}
\label{fig:syn_data_b}
}
}
\caption{\small {\bf (a)} Results averaged over 5 runs for each $\kappa$=50,100,150. vMF centers case is when the true centers are used to generate the code. Empty clusters centers are removed during training for SPKM (hence average number of neurons for SPKM 2 is slightly lower than for AHL $K_w\!=\!2$). {\bf (b)} Correlations between weight vectors (sorted). There are 14 weight vectors, hence $^{14}C_2\!=\!91$ correlations. SPKM and AHL $K_w\!=\!1$ (with pruning off) result in similar levels of correlation.}
\end{figure}

Numerical results are reported in Fig. \ref{fig:syn_data_b} and \ref{fig:syn_data_a}. For each value of $\kappa$, we average over 5 runs (different random directions for the vMF mixture). We clearly see the higher representational power of AHL: the entropies of both the AHL ($K_w\!=\!1,2$) algorithms are higher than their respective counterparts, SPKM 1 and 2, while having similar reconstruction errors. In particular, for $K_w\!=\!2$, we recruit a slightly larger number of neurons, but the codes are far more distributed (higher entropy) while giving slightly better reconstruction error at low $\kappa$.
AHL also results in more decorrelated weights, as seen in Fig. \ref{fig:syn_data_b}, where correlations are plotted for a single run that had 14 centers.

\section{Unsupervised Image Feature Extraction}
\label{sec:ahl_results}

\begin{figure}[h]
\centering
\includegraphics[width=0.9\columnwidth, scale=0.8]{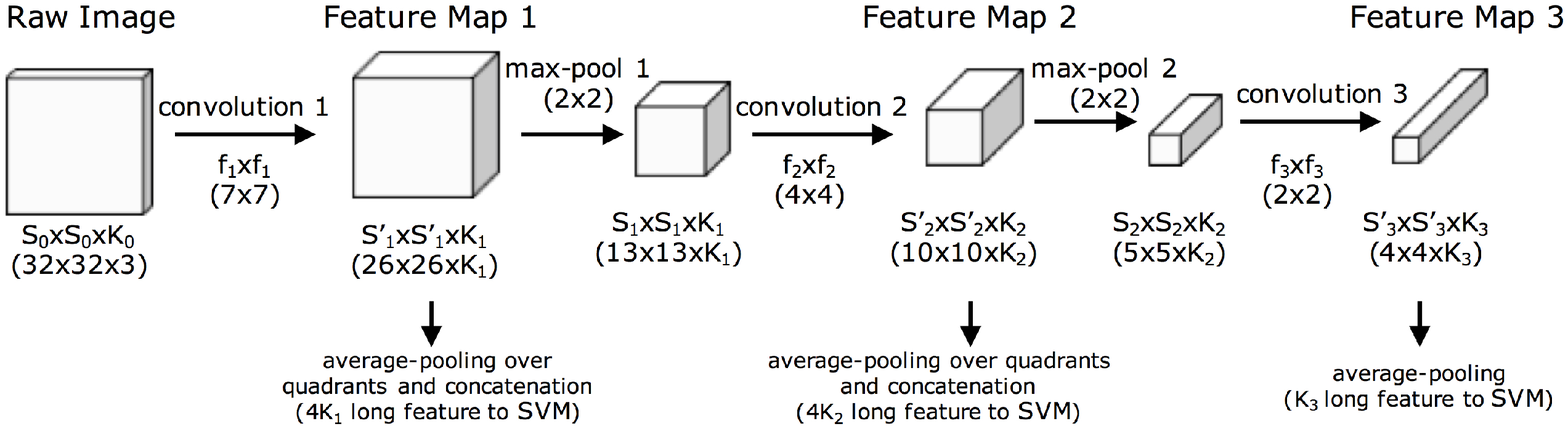}
\caption{\small CNN Architecture. All convolutions are performed with stride 1 and no zero padding, thus $S_i'=S_{i-1}-f_i+1$. Values used in simulations with CIFAR shown in brackets.}
\label{fig:cnn_architecture}
\end{figure}

\subsection{Architecture} 
We use a standard CNN architecture as shown in Fig. \ref{fig:cnn_architecture}. There are 3 convolutional layers interspersed with max-pooling layers. Hebbian learning is used to train the weights and biases of these layers. Layers are sequentially learnt from 1 to 3 in an unsupervised manner. Training data at each layer comprises of $N$ (typically $100-200K$) normalized activations randomly sampled from the lower layer. The 3 intermediary feature maps (Fig. \ref{fig:cnn_architecture}) are used (via average pooling and concatenation) to construct the final features. To study the quality of representations, these features are used to train a linear SVM for classification. At the first layer, raw image patches are processed as described in \cite{coates2011analysis}. 

\subsection{Parameters}
Most of the AHL parameters are set 
to the typical values shown in \ref{alg:ahl_alg}, without optimizing across datasets and layers.  The only parameter that is varied is $A_T$,
which acts a proxy for $K$, the number of hidden units: for fixed $\rho_T$, increasing $A_T$ results in increased $K$ (although $K$
saturates after a point due to the $\rho_U$ pruning constraint). In our experiments, we choose $A_T$ so as to get gradually increasing number of features for higher layers (in the order of a few hundreds), and so that $K_i$ are comparable for the two cases we simulate: WTA ($K_w\!=\!1$) and soft-WTA ($K_w\!=\!3$). The filter and pooling sizes depend on the dataset, and are specified next.

\begin{figure}[h]
\centering
\subfigure[]{\raisebox{3mm}{
\includegraphics[trim=60 50 0 40,clip,width=0.45\columnwidth, scale=0.5]{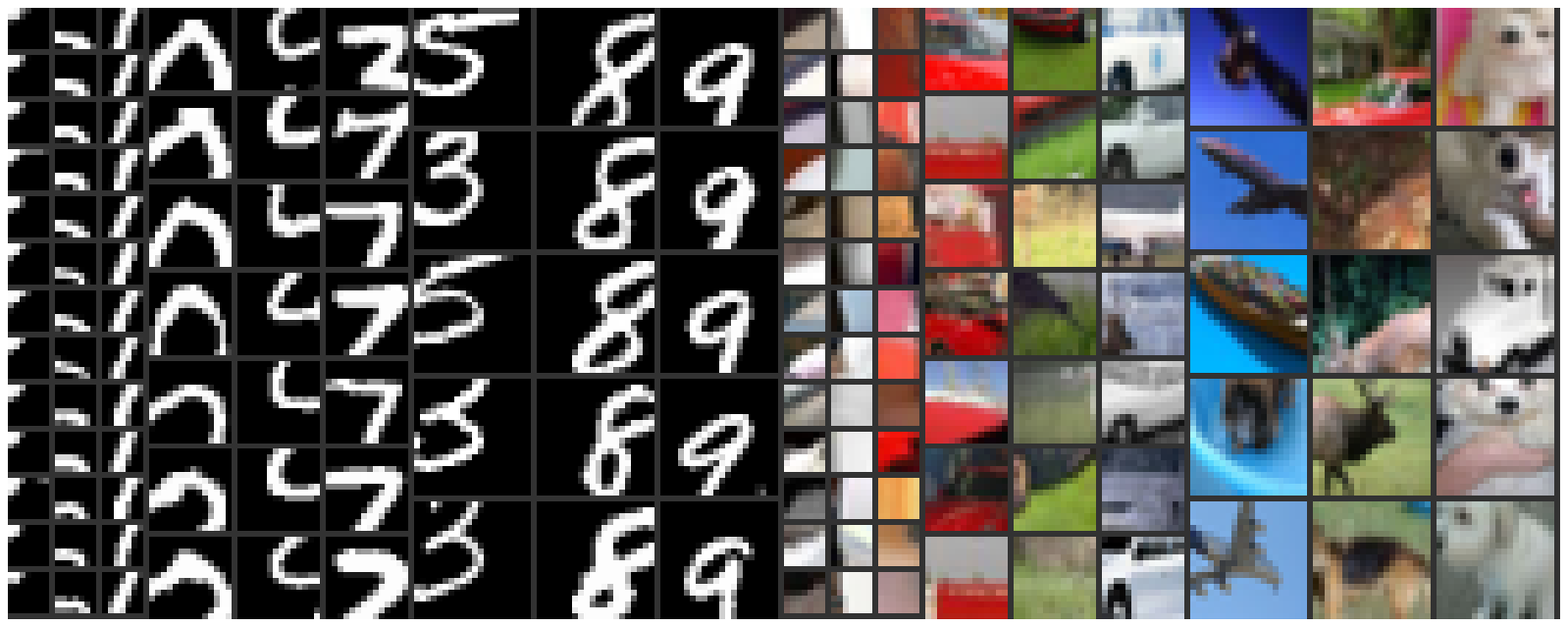}
\label{fig:patches}
}
}
\subfigure[]{\raisebox{2mm}{
\includegraphics[trim=60 5 10 5, width=0.17\columnwidth,scale=0.17]{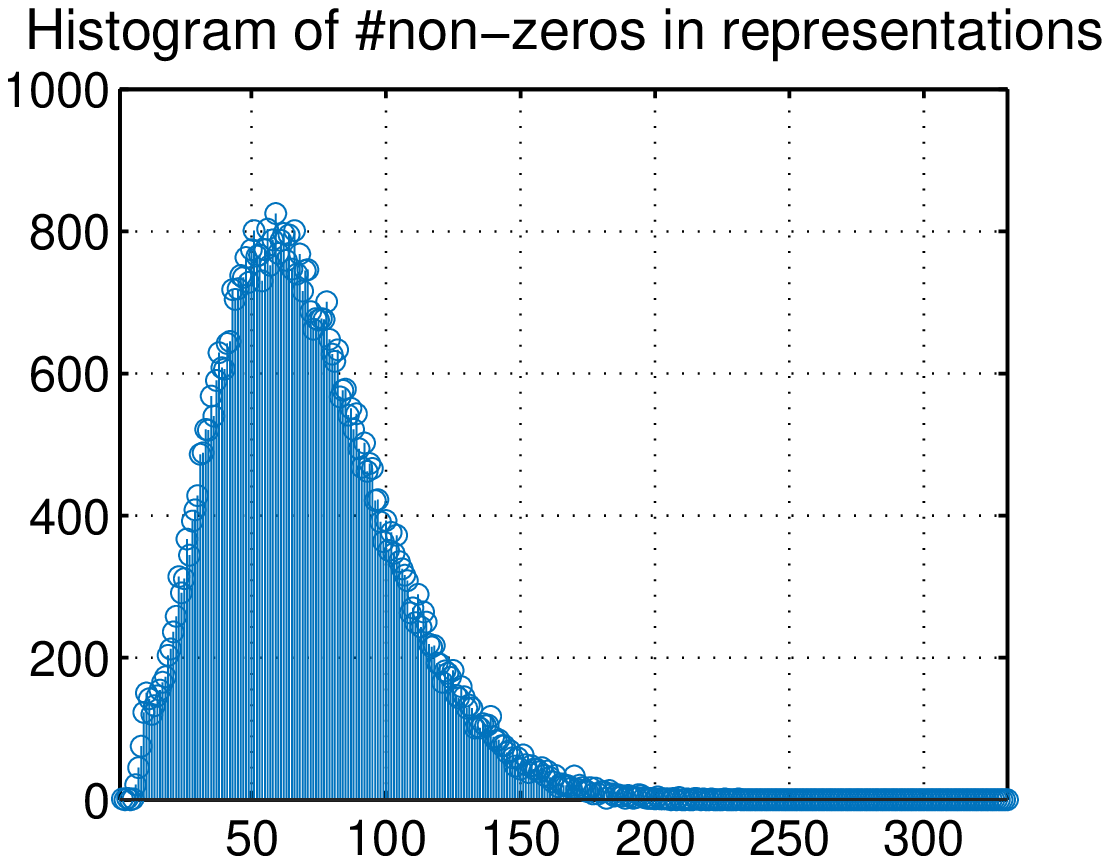}
\label{fig:histogram}
}
}
\subfigure[]{\raisebox{2mm}{
\includegraphics[trim=10 5 60 5, width=0.17\columnwidth,scale=0.17]{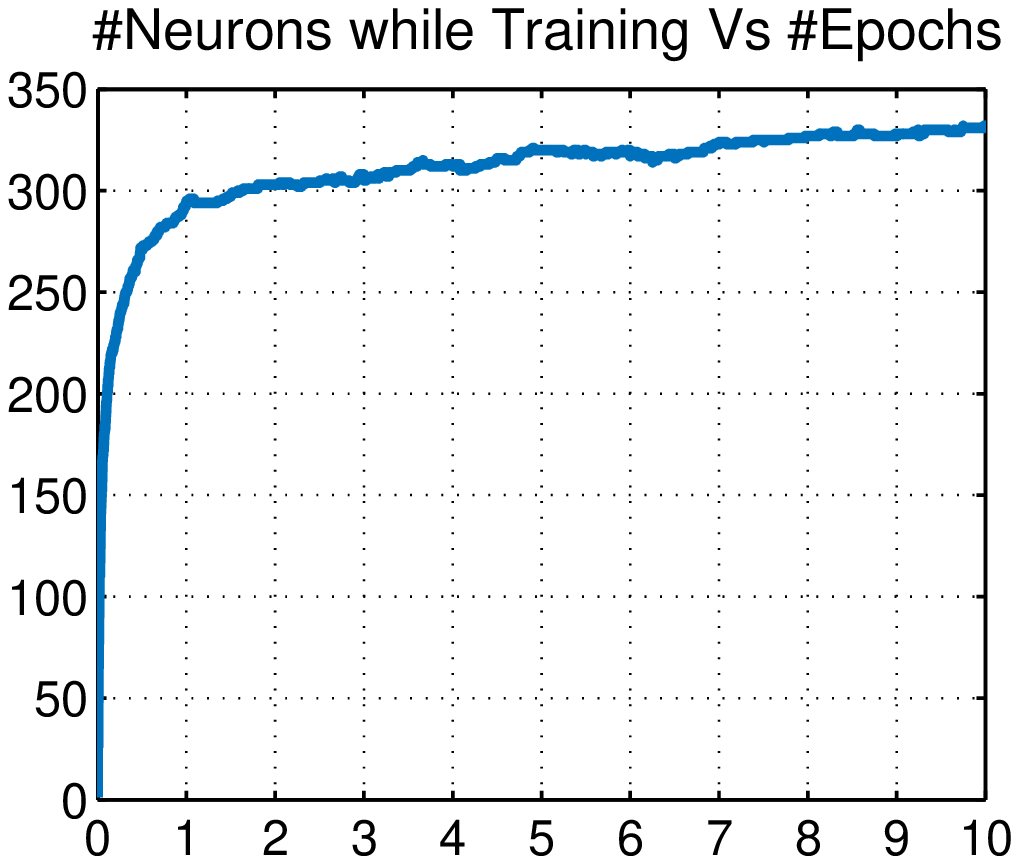}
\label{fig:number_neurons}
}
}
\caption{\small {\bf (a)} Visualization: MNIST and CIFAR top activating patches (corresponding to neurons randomly selected) are shown along the columns (first 3 columns: neurons from layer 1 of MNIST, next 3 columns: layer 2 of MNIST and so on). Filters learnt using AHL, $K_w\!=\!3$ {\bf (b)} Histogram of number of non-zeros neurons in the activation code (total length is 330) {\bf (c)} K, length of ${\bf y}$ i.e. number of neurons recruited as the training progresses (x-axis is number of epochs) (Training using AHL $K_w\!=\!3$, layer 2 CIFAR)}
\end{figure}

\begin{table}[t]
\caption{\small Classification Error rates using unsupervised feature extraction followed by a linear SVM. `1+2+3' denotes concatenation of features from the first 3 layers.
For MNIST, $K_1\!\!=\!\!70,K_2\!\!=\!\!150,K_3\!\!=\!\!260$. For CIFAR, $K_1\!\!=\!\!140,K_2\!\!=\!\!330,K_3\!\!=\!\!600$. For NORB, $K_1\!\!=\!\!170,K_2\!\!=\!\!330,K_3\!\!=\!\!680$. SVM input feature dimension for the 4 cases are: $4K_1,4K_2,K_3,4K_1+4K_2+K_3$}
\label{tab:unsupervised_results}
\centering
\setlength\tabcolsep{3pt}
\resizebox{\linewidth}{!}{%
\begin{tabular}{|c|c|c|c|c||c|c|c|c||c|c|c|c|}
\hline
& \multicolumn{4}{|c||}{MNIST} & \multicolumn{4}{|c||}{CIFAR} & \multicolumn{4}{|c|}{NORB} \\
\hline
\multirow{2}{*}{Layer} & \multirow{2}{*}{SPKM} & AHL & AHL & \multirow{2}{*}{SAE} & \multirow{2}{*}{SPKM} & AHL & AHL & \multirow{2}{*}{SAE} & \multirow{2}{*}{SPKM} & AHL & AHL & \multirow{2}{*}{SAE} \\
& & $\tiny (K_w\!\!=\!\!1)$ & $(K_w\!\!=\!\!3)$ & & & $(K_w\!\!=\!\!1)$ & $(K_w\!\!=\!\!3)$ & & & $(K_w\!\!=\!\!1)$ & $(K_w\!\!=\!\!3)$ & \\
\hline
1 & $1.44\%$ & $1.30\%$ & $\bf 1.26\%$ & $1.47\%$ & $33.88\%$ & $34.42\%$ & $\bf 32.95\%$ & $35.76\%$ & $6.08\%$ & $4.58\%$ & $\bf 3.97\%$ & $6.08\%$ \\
\hline 
2 & $0.98\%$ & $1.00\%$ & $\bf 0.76\%$ & $0.99\%$ & $32.14\%$ & $29.93\%$ & $\bf 28.79\%$ & $35.30\%$ & $6.07\%$ & $4.73\%$ & $4.61\%$ & $\bf 4.35\%$ \\
\hline 
3 & $1.89\%$ & $1.70\%$ & $1.48\%$ & $\bf 1.46\%$  & $44.33\%$ & $38.98\%$ & $\bf 37.05\%$ & $42.65\%$ & $10.56\%$ & $11.46\%$ & $\bf 7.47\%$ & $13.75\%$ \\
\hline
1+2+3 & $0.81\%$ & $0.70\%$ & $\bf 0.65\%$ & $0.72\%$ & $27.49\%$ & $25.46\%$ & $\bf 24.13\%$ & $28.86\%$ & $4.85\%$ & $3.87\%$ & $3.48\%$ & $\bf 2.80\%$ \\ \hline
\end{tabular}
}
\end{table}

\subsection{Datasets}

{\bf MNIST} \cite{lecun1998gradient} is a 10-digit database with 28x28 binary images. We use filter sizes $\{f_1,f_2,f_3\}=\{7,4,2\}$ and $2\times 2$ max-pooling. The size of the receptive field of a neuron onto the raw image is calculated by back-projecting its receptive field to the layer below, and so on, until we reach
layer 0 (the input image). In this case, the sizes are $7\times 7$, $14\times 14$ and $20\times 20$ pixels, respectively, for neurons in hidden layers 1, 2 and 3. One simple way to ``visualize a neuron'' is to display the patches that activate it the most. 
In Fig. \ref{fig:patches} we show such top-5 activating patches for randomly picked neurons in layers 1, 2 and 3. It is intuitively pleasing
to see neurons sensitive to simple edges in layer 1, and to combination of edges in higher layers. {\bf CIFAR} \cite{krizhevsky2009learning} is a dataset of tiny $32\times 32$ color images belonging to 10 classes (e.g., dogs, frogs, ships). The filter sizes used are same as MNIST, also shown in the Fig. \ref{fig:cnn_architecture}. In Fig. \ref{fig:patches} we note that in addition to edges, color sensitive neurons are also learnt. {\bf NORB} \cite{lecun2004learning}, uniform-normalized, is a synthetic dataset of 5 classes of toys photographed under varying lighting and azimuth conditions. The original dataset has $96\times 96$ pixels large grayscale binocular images (hence $K_0=2$). However a large part of the margins are plain background, hence for faster processing we prune the images to $80\times 80$. We use filter sizes $\{f_1,f_2,f_3\}=\{11,4,3\}$ and $3\times 3$ max-pooling which give receptive fields of sizes $11\times 11,\;22\times 22,\;46\times 46$ pixels of raw images patches.

\subsection{Results} 
The distribution of number of active units in the codes, as shown in Fig. \ref{fig:histogram}, is what is expected from sparse distributed codes. Classification error rates are reported in Table \ref{tab:unsupervised_results}. We note that AHL gives better results than spherical K-means (SPKM). We note that $K_w\!=\!3$ performs consistently better than $K_w\!=\!1$, which shows that more distributed representations indeed help.
 
We replace AHL by an autoencoder in each layer in order to compare against both denoising autoencoders (DAE) \cite{vincent2008extracting} (hyperparameters: batch size, noise level, learning rate) and sparse autoencoders (SAE) \cite{ng2011sparse,goodfellow2009measuring} (hyperparameters: weight decay, sparsity penalty coefficient, target activation). The number of hidden units is set to be comparable to the ones reported by the AHL runs. Autoencoder training generally takes much longer, as two layers are trained concurrently, and it was observed that several iterations (of the order of few hundreds) need to be run to get good performance. For the results reported in Table \ref{tab:unsupervised_results}, autoencoder cost function was minimized using 
500 iterations of the quasi-newton L-BFGS method (Matlab minFunc package) and 15 epochs were used for AHL. The time taken, for example, in CIFAR layer 2, was $12.3$ and $74.8$ minutes respectively for AHL and sparse AE (SPKM took about 3 minutes with 5 different randomly initialized runs). This also makes the process of hyperparameter cross-validation for autoencoders very time consuming, which seems to be crucial to get good performance. Although classification performance is not a decisive measure of performance for unsupervised schemes, we found that AHL, while faster and with minimal requirements of tuning, performs consistently better than DAE (which is therefore omitted from Table \ref{tab:unsupervised_results}) and comparable or better than SAE, for the CNN architecture considered here.  We note that SAE features yield slightly better performance for NORB when concatenating all three layers, but performance with CIFAR is much worse. 
 
The error rates we obtain are comparable to some of the best rates reported in literature using unsupervised features for classification without model averaging and data augmentation, such as, MNIST: $0.82\%$ using deep belief networks \cite{lee2009convolutional}, $0.64\%$ \cite{ranzato2007unsupervised} using unsupervised features fed to a supervised two layer NN, NORB: $2.52\%$ \cite{akbas2014framework} using finer pooling (over 5x5 regions instead of quadrants), CIFAR: $80\%$ \cite{coates2011analysis}, $82\%$ \cite{coates2011selecting}, $80.1\%$ \cite{makhzani2014winner} accuracy, using large number of feature maps ($>4k$). Our focus in this work is not to finely tune or over build the networks to beat state of the art benchmarks, but to highlight a fast method which requires minimal tuning  and performs well with reasonable network size. We note that our performance improves further by incorporating some of the techniques reported in earlier work, giving error rates of $2-3\%$ for NORB by finer pooling as in \cite{akbas2014framework}, and accuracy of up to $80\%$ for CIFAR by increasing $K_1$ as in \cite{coates2011analysis}. 

Note that classification accuracies deteriorate considerably when using layer 3 alone. This points out the importance of incorporating labels while training higher layers for classification tasks, so that more discriminative features could be learned. A promising idea that we are currently exploring is to combine Hebbian and anti-Hebbian mechanisms to generate class-specific neurons. Results will be reported in future publications.

\section{Conclusions}
\label{sec:conclusions} 

We have demonstrated that the Hebbian principle, with appropriate competitive mechanisms, provides a powerful basis for designing
unsupervised learning algorithms.  The AHL algorithm presented here is a radical departure from prior approaches: instead of trying to
minimize a cost function, AHL is able to directly target desirable properties such as sparsity and decorrelation using neuroplausible mechanisms.
The training complexity of AHL is less than that of autoencoders, while the features obtained perform better in the experiments considered here.
To the best of our knowledge, this is the first paper in which learning abstractions firmly grounded in neuroscientific models have been demonstrated to be competitive
with modern machine learning techniques. We hope that the promising results reported here stimulate further investigation into bridging the gap between neuroscience
and machine learning, with the goal of enhancing our understanding in each area.

As with any unsupervised approach, as we go up the layers, there is a tendency to ``waste'' neurons on modeling features that are not ultimately
informative (e.g., if classification is our end goal, modeling different types of backgrounds may not be useful). An exciting area for future work is to explore semi-supervised Hebbian learning techniques which can exploit high volumes of unlabeled data, while employing small amounts of labeled data to prune such task-agnostic neurons.

\bibliographystyle{ieee}
\bibliography{ref}

\end{document}